\documentclass{article}
\usepackage{ijcai22}

\usepackage{times}
\usepackage{soul}
\usepackage{url}
\usepackage[hidelinks]{hyperref}
\usepackage[utf8]{inputenc}
\usepackage[small]{caption}
\usepackage{graphicx}
\usepackage{amsmath}
\usepackage{amsthm}
\usepackage{booktabs}
\usepackage{algorithm}
\usepackage{algorithmic}
\title{Threshold Designer Adaptation: Improved Adaptation for Designers in Co-creative Systems}

\author{
Emily Halina
\And
Matthew Guzdial
\affiliations
Department of Computing Science, Alberta Machine Intelligence Institute (Amii)\\
University of Alberta, Edmonton, Alberta, Canada\\
\emails
\{ehalina, guzdial\}@ualberta.ca
}

\begin{document}

\maketitle

\begin{abstract}




To best assist human designers with different styles, Machine Learning (ML) systems need to be able to adapt to them.
However, there has been relatively little prior work on how and when to best adapt an ML system to a co-designer.
In this paper we present threshold designer adaptation: a novel method for adapting a creative ML model to an individual designer.
We evaluate our approach with a human subject study using a co-creative rhythm game design tool.
We find that designers prefer our proposed method and produce higher quality content in comparison to an existing baseline.
\end{abstract}

\section{Introduction} 




The study of creative Artificial Intelligence (AI) has increased in recent years \cite{veale2019computational}.
Creative AI agents have been proposed and developed for many domains, including music, painting, and robotics \cite{carnovalini2020computational,yalccin2020empathic,fitzgerald2017human}.
These advancements have also led to the development of systems that allow human designers to interact directly with AI agents co-creatively \cite{guzdial2018co}.
Many of these co-creative AI systems utilize modern Machine Learning (ML) methods due to the recent strides ML methods have made in a number of creative fields \cite{franceschelli2021creativity}. 

A major challenge in designing ML models for co-creative systems is ``adaptability,'' the notion of changing a model's behaviour to fit a given designer's preferences.
Particularly within artistic domains, there is a need for co-creative systems to adapt to an individual designer's artistic sensibilities.
For example, two artists who interact with the same co-creative art tool may have vastly disparate styles such that an ML model could not match both of them without adaptation \cite{gatys2015neural}.
However, it is still not fully understood how and when it is best to adapt an ML model to a designer within a co-creative context.
Further understanding this problem of designer adaptation would allow us to bolster the capabilities of these systems in a variety of domains.



In this paper we explore the problem of designer adaptation within the domain of game design.
AI has long been used to aid game developers with both game design and asset creation tasks \cite{yannakakis2018artificial}.
In particular, the application of AI methods to generate game content is known as Procedural Content Generation (PCG) \cite{hendrikx2013procedural}.
Several co-creative PCG systems---with and without the use of ML---have been developed in prior work \cite{guzdial2018co,charity2020baba}.
The application of ML to co-creative PCG is relatively novel, and as such there has not been deep prior investigation into adaptation for human designers \cite{guzdial2019friend}.



Towards the goal of effectively adapting to individual designers, we propose ``threshold designer adaptation,'' a novel method for adapting creative ML models to individual designer input.
This strategy revolves around determining a threshold of training data to retrain at by approximating human interaction using existing data.
We hypothesize that this strategy will help models adapt to individual designers in a meaningful way, particularly in domains that require internal consistency and structure, such as painting and music.
This is due to the model retraining only after there is enough context for the human-authored data to be meaningful.

In order to investigate the problem of designer adaptation, we designed and developed KiaiTime \cite{halina2021kiai}, an adaptive co-creative PCG system for the rhythm game Taiko no Tatsujin (Taiko) \cite{tpgpl_semro_megaapplepi_2020}.
The purpose of this tool is to help a designer create Taiko charts: a series of pre-determined game objects (notes) that are hit by the player to the beat of a song.
An example of a chart is shown in Figure \ref{fig:interface}.
Given that expert Taiko designers regularly collaborate in chart generation, we contend this domain is well-suited for studying co-creation and designer adaptation.
KiaiTime uses an ML model to determine where to place notes based on both the song's audio and previously placed notes.
This ML model is set up to test different adaptation strategies, with KiaiTime collecting training data from individual designers as they interact with the system.

To evaluate the effectiveness of threshold designer adaptation, we conducted a human subject study using the KiaiTime system.
We found that designers preferred interacting with the model using threshold designer adaptation over a baseline approach. 
We also found that designers created higher quality charts when interacting with a model using threshold designer adaptation. 



The main contributions of this work to the broader creative AI community are as follows:
\begin{itemize}
    \item Threshold designer adaptation, a novel method for adapting co-creative ML systems to designers.
    \item A discussion and system overview of KiaiTime, a co-creative AI system for studying designer adaptation.
    \item A detailed account of the methodology, design, and results of a human subject study used to evaluate a co-creative AI system.
\end{itemize}



\section{Related Work}



\subsection{AI Co-creation}

Human-AI co-creation systems are systems in which a human designer and AI agent work together on a task \cite{davis2013human}.
Many co-creative AI systems generate content based on a given ``prompt'' from a human user \cite{galanosaffectgan,gero2019metaphoria,champandard2016semantic}.
However, these systems only allow for minimal collaboration and adaptation, as the human only influences the input given to an unchanging AI model.
In comparison, we focus on iterative, long-term interactions that change the underlying model's predictions to appropriately suit a user on arbitrary input.
Other co-creative AI systems act as enhanced content ``editors'' that supplement human creators in various ways, falling under the broader category of creativity support tools \cite{frich2019mapping}.
An example is MakeWrite, a co-creative system for helping users with communication disorders write via the use of automatic constraints \cite{neate2019empowering}.
These systems focus on a human user adapting their behaviour toward a desired outcome rather than an AI adapting to the user.

Our framework is a mixed-initiative co-creative system, which is a type of system in which both the AI agent and human user ``take the initiative'' contributing to a collective artifact \cite{liapis2016mixed}.
A distinguishing quality of mixed-initiative systems is the shared agency between the human and AI agent \cite{zhu2018explainable}.
However, existing mixed-initiative AI systems generally do not retrain at all, leaving the AI partner to be ``stubborn'' and unchanging.

\subsection{Games}
Co-creative frameworks utilizing PCG techniques have been developed for many game domains.
The majority of prior mixed-initiative and co-creative PCG approaches do not utilize ML techniques, and instead rely on other non-ML AI approaches \cite{sturtevant2020demonstration,charity2020baba,liapis2013sentient}.
While some of these tools do adapt to designers, they do so only in a human-authored and predetermined way, limited to their specific game domain.
We instead investigate general adaptation strategies.

Procedural Content Generation via Machine Learning (PCGML) refers to the subsection of PCG approaches utilizing ML.
There are still many open problems to be explored within co-creative PCGML systems, one of which is designer adaptation \cite{summerville2018procedural}.
Due to intrinsic challenges with ML approaches such as lack of training data for individual users, the majority of prior co-creative PCGML systems for games do not include designer adaptation.
The singular example of an adaptive co-creative PCGML system evaluated with a human subject study is Guzdial et. al.'s ``Morai Maker'' \cite{guzdial2019friend}, a mixed-initiative PCGML editor for Super Mario Bros.-style levels.
Within the Morai Maker editor, the AI agent actively trains based on implicit user feedback that takes into account whether the AI's suggestions were kept or deleted by the human partner.
This training methodology informed the ``naive'' training strategy which was used as a baseline in our human subject study.
To the best of our knowledge, our system KiaiTime is the first co-creative tool for chart generation.


\begin{figure*}[h]
    \centering
    \includegraphics[scale=0.33]{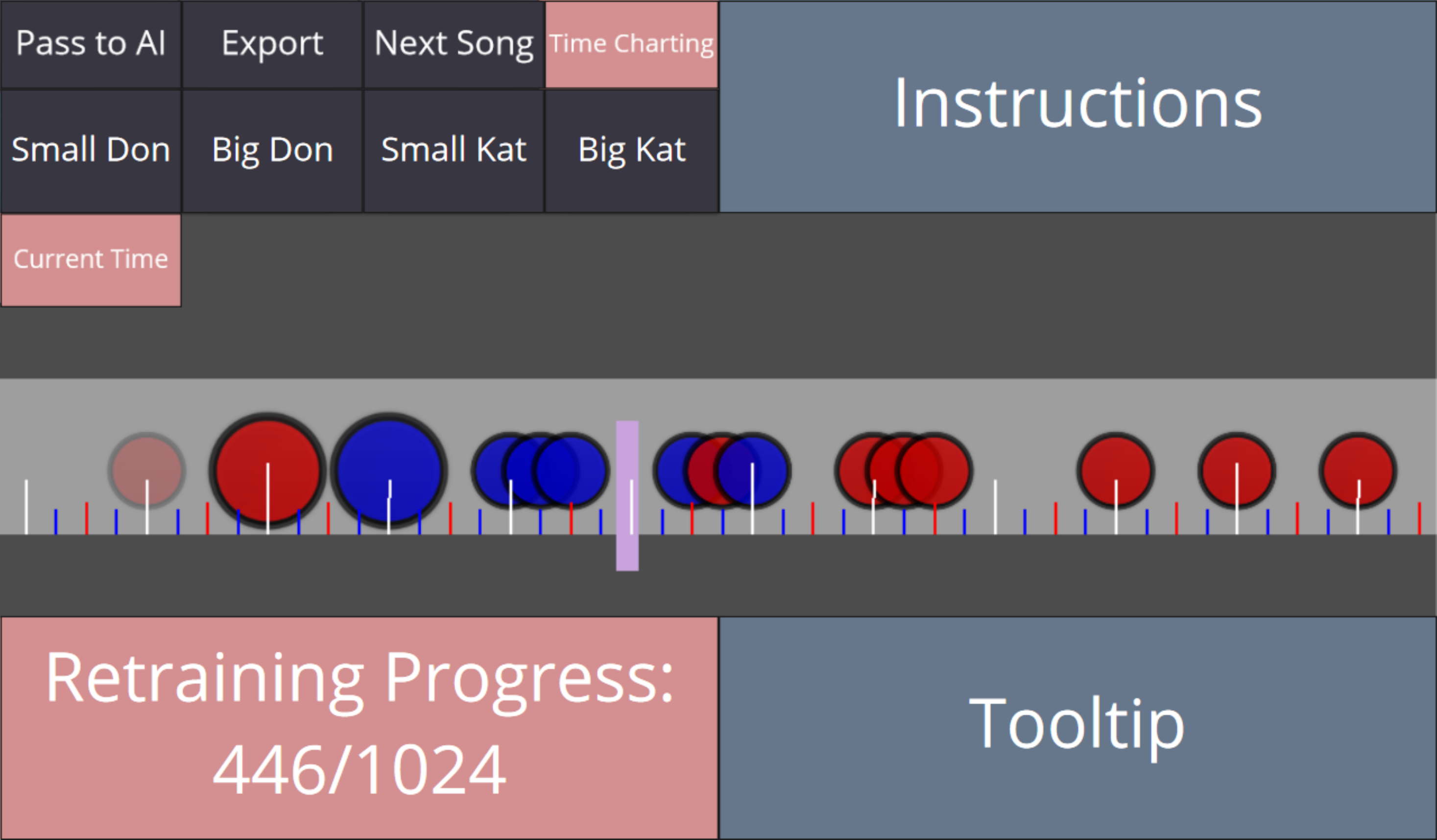}
    \caption{Screenshot of the final KiaiTime editor. Users place notes on the grey timeline in the center of the screen to create a rhythm game chart along with the AI partner. Users move the timeline in real time to the beat of the song they are charting.}
    \label{fig:interface}
\end{figure*}

\section{Threshold Designer Adaptation}


Our primary contribution is our threshold designer adaptation approach. 
Threshold designer adaptation is a novel approach for adapting creative ML models to individual designers within co-creative tools.
This approach relies on determining a threshold $\delta$ of training data needed to begin retraining, thus the name. 
To determine the value of $\delta$ we rely on a simulation of co-creative interaction with existing human data.
In this way, we can approximate how different values of $\delta$ will perform when a human designer interacts with the AI.






There is high variability in the amount of retraining data needed to make a meaningful change to an ML model's output depending on architecture and domain.
As such, we propose $\delta$ be determined empirically by employing existing human data to find the best value to retrain at.
In order to do this, we mimic the co-creative process as closely as possible, using a collection of existing human data created by different designers.
For each designer, the initial ML model is retrained on the first $\delta$ training instances created by that designer for a range of possible $\delta$ values.
After this retraining, we measure the model's performance on the remaining training instances created by that designer.
This is done by measuring how well the model matches the designer's unseen training instances via a domain-dependant process, note accuracy in our case.
Values of $\delta$ that perform well theoretically produce models which are further adapted to the individual designer using them.
Given the lack of prior co-creative PCGML work that features designer adaptation, to the best of our knowledge this methodology for determining ``when to adapt'' to a given designer in a co-creative context is novel.

At runtime, we begin with a co-creative system using an ML model that has been pretrained on training data that reflects the output of many different designers.
As an individual designer interacts with the co-creative system, the system collects ``retraining data'' $D_r$ based on the input of that designer.
The form of this input varies depending on application, and in our case consists of  segments of designer-created Taiko chart.
After sufficient retraining data is collected, the ML model is retrained in order to tweak its output to accurately match the designer using the tool.

Whenever a designer's input creates a new instance of training data $(x, t)$, the system adds it to the set of retraining data $D_r$. When this occurs, the following inequality is evaluated:
\begin{align*}
    |D_r = \{x^{(m)}, t^{(m)}\}_{m=1}^M| \geq \delta (k + 1)
\end{align*}
If the total number of retraining samples $M$ is greater than the predefined retraining threshold $\delta$ multiplied by the number of times we have retrained $k$, the system retrains the ML model.
Each time the system retrains the model, $k$ increments.


\section{KiaiTime}

In this section we provide an overview of KiaiTime, a co-creative system for rhythm game chart generation.
KiaiTime utilizes threshold designer adaptation to adapt its ML model to act as a partner for human designers.
Section 4.1 reviews the details of KiaiTime's ML model, along with our methodology for determining the value of the retraining threshold $\delta$ for our specific task.

The KiaiTime editor is a co-creative PCGML system we used as a testbed for different designer adaptation techniques \cite{halina2021kiai}.
KiaiTime's name alludes to ``kiai time,'' a mechanic in Taiko and other rhythm games which provides a visual highlight to the most intense parts of songs.
A video showcasing the functionality of the KiaiTime editor is included within the supplementary materials,\footnote{For supplementary materials and source code, see \url{https://github.com/emily-halina/KiaiTime}} along with further descriptive detail regarding the editor.


Figure \ref{fig:interface} depicts a screenshot of the final KiaiTime editor.
In the editor, human users collaborate with an AI partner in a turn-based structure to create a Taiko chart for a pre-selected song.
Taiko is a rhythm game where players attempt to hit a series of pre-determined game objects (notes) in time to the beat of a song.
KiaiTime allows the user to create this series of notes, which are called ``charts.''
These charts are later playable in real time.

Designers have access to the four different types of Taiko notes, each representing a different button press or ``drum hit.''
Designers place notes on the timeline using left mouse clicks, and remove notes using right mouse clicks.
The multi-coloured ticks on the timeline each represent a 1/16th beat of a given song, with placed notes automatically snapping to these ticks to match the music.
These ticks are coded by size and colour, representing the beat division of different measures.
This type of colour-coded beat division is standard within other Taiko editors \cite{tpgpl_semro_megaapplepi_2020}.

At any point the user may press the ``Pass to AI'' button to begin the process of ending their turn.
Once the button is pressed, the user selects a region for the AI partner to ``fill in,'' querying the ML model for its predictions across the given region.
The ML model uses the audio and chart data around the selected region to make its predictions, outputting a series of notes that the editor places in the chosen region.


\subsection{Threshold Designer Adaptation Implementation}

KiaiTime uses the pre-existing ``TaikoNation'' system for generating the AI partner's predictions \cite{halina2021taikonation}.
TaikoNation utilizes a Long Short Term Memory Recurrent Neural Network (LSTM RNN) to generate predictions of where to place notes in a Taiko chart based on both the song being charted as well as previously placed notes.
The model is pretrained on 100 high-quality human-authored charts from a variety of designers, and is retrained on the fly when used in KiaiTime to match the style of individual designers.
These 100 charts are split into 23 millisecond segments, resulting in approximately 1.3 million training samples.
This model was chosen over other rhythm game chart generation ML models \cite{donahue2017dance,lin2019generationmania} due to TaikoNation's improved notion of human-like ``patterning,'' the placement of game objects in a congruent fashion in terms of both rhythm and chart context.
The full technical details regarding TaikoNation's architecture and training data can be found within the corresponding TaikoNation paper \cite{halina2021taikonation}.

In order for KiaiTime to best adapt to human designers using threshold designer adaptation, we needed to determine both how and when to adapt the pretrained TaikoNation model to an individual user.
This requires us to determine both the retraining threshold $\delta$ described in Section 3, as well as other hyperparameters which affect the model's learning process.
To do this, we utilized a grid search approach, exhaustively sweeping over a manually defined space of hyperparameters in order to find the most effective combination \cite{feurer2019hyperparameter}.
The hyperparameters we tuned in addition to the retraining threshold $\delta$ were learning rate $\alpha$, maximum number of training epochs $\epsilon$, and batch size $b$.
For each combination of hyperparameters, we trained the model on the first $\delta$ training instances extracted from a chart, then queried the model to complete the rest.
The model's output was judged both on the diversity of patterns used within a chart, as well as how closely the model was able to match what the human author actually created.
We repeated this process for three charts made by three different designers, attempting to recreate the co-creative environment as closely as possible.
We chose these three charts due to the disparate artistic sensibilities of their respective designers.
The system converged to a consistent $\delta$ value for all three designers, and so we felt no need to expand to additional designers.
For each hyperparameter, we selected a small set of prospective values to test with the goal of shrinking the search space and preventing overfitting.
The choices for the retraining threshold $\delta$ correspond to $10\%, 20\%, 40\%,$ and $80\%$ of a 2-minute chart respectively.

Through the grid search, we found the following set of hyperparameters that were used for retraining TaikoNation.
\begin{align*}
    \{ \delta = 1024 \hphantom{1} (20\%), \hphantom{1} \alpha = 10^{-3}, \hphantom{1} \epsilon = 5, \hphantom{1} b = 4 \}
\end{align*}
This set of parameters performed well on predetermined metrics from prior work \cite{halina2021taikonation} while also producing impressive qualitative results from our perspective. 
For consistency, we used these hyperparameters for both threshold designer adaptation and our baseline approach discussed in Section 5.1.

\section{Human Subject Study}

In order to evaluate the performance of threshold designer adaptation, we conducted a human subject study using KiaiTime.
During our study design, we identified 2 major objectives which informed our study methodology and our interpretation of the study's results. These objectives are as follows:
\begin{itemize}
    \item \textbf{(O1):} To identify whether threshold designer adaptation helps to adapt ML models to designers.
    \item \textbf{(O2):} To further understand the open problem of adapting to designers within a co-creative context.
\end{itemize}

O1 arose from the desire to evaluate the value of threshold designer adaptation as an approach.
O2 arose from the desire to inform future research into co-creative adaptation.


In order to complete these objectives, we conducted a within-subject study with 20 human designers, comparing threshold designer adaptation to an existing naive baseline approach defined in Section 5.1.
We chose a within-subject structure rather than a between-subject structure due to the vast differences between individual designers.
By comparing the results of each designer with each adaptation strategy, we can evaluate which approach helps our ML model adapt best to individual designers.


Prior to conducting the study, we identified the following two hypotheses:
\begin{itemize}
    \item \textbf{(H1):} Threshold designer adaptation will help designers create better charts than the baseline naive approach.
    \item \textbf{(H2):} Designers will prefer threshold designer adaptation to the baseline naive approach.
\end{itemize}



\subsection{Naive Adaptation Strategy}
Our baseline adaptation strategy, which we refer to as the ``naive'' strategy, entails retraining the model whenever new training data is generated by the human user.
In contrast to threshold designer adaptation, whenever a new piece of data $(x, t)$ is added to the retraining dataset $D_r$, the ML model is retrained using all of the data in $D_r$.
This naive strategy is analogous to the implicit training strategy used in the Morai Maker system \cite{guzdial2019friend}.
We chose the adaptation strategy from the Morai Maker system as, despite the difference in genre, it is the only other adaptive PCGML tool for a game design task.
Thus, this naive approach represents the current state of the art for this task.
This training is done with the same set of empirically determined hyperparameters used by our threshold designer adaptation strategy.
We also use the same TaikoNation pre-trained model as the base model, which is retrained with individual designer's data \cite{halina2021taikonation}.
We made these decisions to allow for consistency in the comparison between the two strategies.

\begin{table*}[h]
\centering
\begin{tabular}{lrrrrr}
\toprule
\textbf{Approach} & \textbf{Avg Time Spent} & \textbf{Avg End Turn Count} & \textbf{Human Notes Kept} & \textbf{AI Notes Kept} & \textbf{Patterns Used}\\
\midrule
\textbf{Naive}     & 20.6 mins           & 39.5                    & 74.2\%                    & 59.6\% & 18.83\%                \\
\textbf{Threshold} & 24.1 mins           & 48.6                    & 72.1\%                    &\textbf{*72.5\%} & \textbf{*21.72\%}             \\
\bottomrule
\end{tabular}
\caption{Quantitative information extracted from participant logs and created charts. Average Time Spent is in minutes, Average End Turn Count is the average number of times the ``Pass to AI'' button was pressed. Human Notes Kept and AI Notes Kept are the average percentages of notes placed by each respective agent that were not deleted. Patterns Used is the average percentage of possible patterns used in charts created with each respective adaptation strategy. Statistically significant results $(p < 0.05)$ are denoted with an asterisk (*).}
\label{all_logs}
\end{table*}

\begin{table*}[h]
\centering
\begin{tabular}{lrrrr}
\toprule
                   \textbf{Approach}& \textbf{Learned Better} & \textbf{Most Frustrating} & \textbf{Higher Quality} & \textbf{Enjoyed Most} \\
\midrule
\textbf{Naive}     & 8                       & 10                        & 8                       & 6                     \\ 
\textbf{Threshold} & 11                      & 9                         & 9                       & \textbf{*11}                    \\
\bottomrule
\end{tabular}
\caption{Quantitative survey results for the naive and threshold adaptation strategies across all participants. Each ranking question was optional, and could be skipped if a participant had no strong preference. Statistically significant results $(p < 0.05)$ are denoted with an asterisk (*)}
\label{all_survey}
\end{table*}

\subsection{Study Methodology}
In this subsection we briefly cover our study methodology in the order a subject would experience it.
Our study design was reviewed and approved by our university ethics board (Pro00112032).
For the full details of our study methodology, please refer to the supplementary materials.

Participants reached out to research personnel via email to book a 1 hour timeslot after seeing publicly posted advertisements for the remote study.
After acknowledging a consent form, during their timeslot participants downloaded a copy of the KiaiTime editor to their local machine.
Throughout the study this editor interfaced with a server run by research personnel which handled the adaptation and contributions of the ML model.

After a brief tutorial explaining the game domain of Taiko and the functionality of the editor and AI partner, participants were tasked with creating two charts for two randomly ordered songs.
These songs were \textit{polygon} by Sota Fujimori and \textit{citrus} by kamome sano, two electronic dance music tracks from rhythm game soundtracks with similar rhythmic structure and length.
While creating each chart, participants had access to the ML model employing either the naive or threshold designer adaptation strategy chosen at random.
When participants moved on to their second song, the ML model was reset to the default model, and the adaptation strategy was changed to the unused strategy.
Both the ordering of the songs and ordering of adaptation strategies were randomized using a study ID system in order to ensure an even distribution of conditions.

We collected information from participants in three ways.
The first was through a survey which asked participants to rank the two approaches in qualitative categories, along with asking demographic questions.
The ranking questions asked participants which ML model learned better from them, was more frustrating to use, created higher quality contributions, and which they enjoyed most.
The survey is based on prior co-creative system surveys, and the full text is provided within the supplementary materials \cite{guzdial2019friend}.
The second was through a logging file which tracked the time participants spent in the editor, the amount of times they passed control over to the AI partner, and the percentage of notes they kept from both the AI and their own contributions.
The third was from the charts the participants produced when working with each adaptation strategy, which we analyzed using a pre-established metric for Taiko charting quality \cite{halina2021taikonation}.

\subsection{Study Results}







\begin{figure}
    \centering
    \includegraphics[scale=0.35]{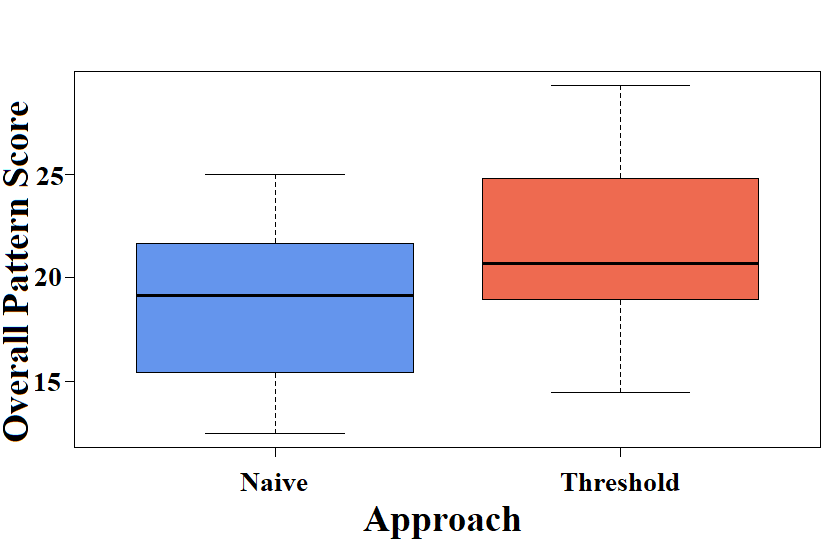}
    \caption{Boxplot depicting the overall pattern scores of charts produced with the naive and threshold adaptation strategies.}
    \label{fig:boxpot}
\end{figure}


In this subsection we discuss the results of our study, and provide an analysis of the quantitative data.
20 participants completed our study and survey.
Of these participants, 17 identified as male, 2 identified as female, and 1 chose to not disclose their gender. 
18 participants placed themselves in the 18-24 age range, with 2 participants placing themselves in the 25-30 age range.
Half the participants were ``experts,'' meaning they have authored high-quality charts according to community standards, and the other half were ``novices'' \cite{tpgpl_semro_megaapplepi_2020}.
This population does not have sufficient diversity to draw broad, universal lessons from this study. 
However, this population is consistent with the general demographic of chart creators, and thus fits our needs for the initial evaluation of threshold designer adaptation.

For each of our demographic categories (age, gender, experience level), we ran a multivariable analysis of variance (MANOVA) test with the survey ranking results and logged data as dependent variables.
We found no significant correlation between any of the demographic categories and the study results.
We also checked if either the ordering of the songs or the order of the adaptation strategies made a significant impact on the results.
We found neither variable had a significant impact on the results, and thus we can safely treat our data as all coming from the same condition.

Table \ref{all_logs} shows the quantitative information extracted from each participant's logs and charts.
We focus on four different pieces of information drawn from the logs divided by adaptation approach: the average time spent charting with the AI partner, the average number of times participants gave the AI partner a turn, and the percentages of notes placed by the human and AI that were kept by the end of the charting process.
We deemed these statistics to be important as they represent the amount of interaction between the participants and AI partners, as well as the quality of that interaction.
Similar analysis of logged files has been used in prior work \cite{guzdial2019friend}.
We found that participants spent more time interacting with the threshold partner, gave the threshold partner a turn more frequently, and kept substantially more notes from the threshold partner in comparison to the naive partner.
To evaluate if there was a significant difference in the distributions of each of these statistics, we performed a Mann-Whitney Wilcoxon test.
We chose this test because we found with the Shapiro-Wilk normality test that our data did not follow a normal distribution $(p < 0.0005)$.
With these tests, we found there was a statistically significant difference between the ratio of AI partner notes kept from the naive and threshold strategies with $p < 0.03$.
This suggests that participants were more accepting of the AI partner's contributions when it was using threshold designer adaptation, and may imply the contributions are of a higher quality than those from the naive strategy.

To further support H1, we analyzed the quality of charts created with both adaptation strategies.
We examined the proportion of possible patterns used in charts created with each adaptation strategy, called Overall Pattern Score. 
Patterns are defined as unique rhythmic sequences which are 8 timesteps (184ms) in length.
This is a suitable metric for estimating chart quality empirically, as higher quality charts tend to use a wider variety of patterns \cite{halina2021taikonation,5argon_2018}.
Figure 2 depicts a boxplot of the Overall Pattern Scores for charts created using each adaptation strategy.
After ensuring that the Overall Pattern Score for both the naive and threshold charts followed a normal distribution with the Shapiro-Wilk normality test (with significance threshold $p<0.05$), we performed a paired t-test on the two distributions.
With this test we found the charts created with the threshold partner used significantly more patterns than the charts created with the naive partner with $p < 0.05,$ 95\% CI $(-\infty, -0.08)$.
These logged results support our hypothesis (H1) that threshold designer adaptation would help designers create better charts than the baseline naive approach.
While further investigation is required, we would encourage other human-AI co-creative systems to adapt with the threshold designer adaptation approach.


Table \ref{all_survey} contains the results of the survey ranking questions across all participants divided by adaptation strategy.
We found an overall positive preference towards the threshold strategy for each question.
To see if there was a significant difference in distribution within any of the survey questions, we performed a Mann-Whitney Wilcoxon test on each of the experiential features.
We chose this test because this data does not follow a normal distribution.
With this test, we found a significant difference between the two adaptation strategies in the ``enjoyed most'' ranking with $p<0.05$.
This result supports our hypothesis H2 that designers would prefer threshold designer adaptation over the baseline naive approach.


\section{Discussion}


In this section we reflect on our study, addressing the limitations of our study design and technical approach, and discuss potential avenues for future work.

\subsection{Limitations}
Our major evaluation limitation is the use of only a single creative domain.
In an ideal evaluation, we would test the approach across multiple creative domains to assess its performance.
We believe our study was sufficient for our initial proposal of the method, leaving the application to other domains as future work.

Another limitation is the lack of a strong assurance that our chosen threshold value $\delta$ is the ``correct'' choice.
With additional resources, we would be able to sweep over a larger number of threshold values, narrowing down to an exact correct choice using a larger set of individual designers.
However, even with only the three charts we used to determine the value of $\delta$, our results suggest our chosen value was at least a reasonable choice.

A potential concern is that retraining in general is a detriment to the ML model's performance.
However, we found during our grid search that models trained using threshold designer adaptation outperformed the baseline TaikoNation \cite{halina2021taikonation} model in recreating a portion of a human chart.
Nevertheless, it is important to consider this possibility in future work on designer adaptation.

\subsection{Future Work}
Threshold designer adaptation has the potential to be applied in many domains for the purpose of co-creativity.
This strategy can be used for any arbitrary set of hyperparameters and metrics for evaluation that are chosen to suit a specific domain.
While we offer an improved strategy for designer adaptation over the prior state of the art approach, there is still no empirically defined ``best practice'' for adapting to users.
The problem of designer adaptation is very domain specific, and thus in order to make progress towards finding this ``best practice,'' it must be evaluated with more co-creative systems in different creative domains.
However, due to the lack of adaptation research and the quality of our results, we view our threshold designer adaptation as the new state of the art for the task of designer adaptation.
One additional avenue for future work is the usage of adaptive co-creative agents on a longer term basis with individual users.
This could entail looking at adaptation strategies that could be applied over weeks or months, examining how creative ML models can be fine-tuned to fit a particular designer over a longer time span.
This could be useful for artistic practices that may benefit from long-term context of a designer's tastes, such as music composition or painting.

As with any artificial intelligence application, it is important to acknowledge the potential risk of the application taking away the livelihood of people or discouraging human creation.
KiaiTime and the threshold designer adaptation strategy are designed with co-creation in mind rather than as a replacement for a human designer.
The systems described in this paper are not a replacement for human designers, but rather function as tools for human designers to use to assist their creative process and inspire novel designs.
We will continue to reflect on these potential risks when developing future versions of the tool and approach.

\section{Conclusions}

In this paper we presented threshold designer adaptation, a novel method of adapting to individual designers in a co-creative context.
We ran a human subject study investigating two strategies for designer adaptation using KiaiTime, the first co-creative PCGML system for rhythm game chart design.
We found that designers made higher quality charts when interacting with an ML model using threshold designer adaptation over a baseline approach.
We also found designers preferred working with the model using threshold designer adaptation.
We hope our approach contributes to the future of designer adaptation in co-creative systems and allows ML models to better fit individual designers.

\section*{Acknowledgements}
This work was funded through a Natural Sciences and Engineering Research Council of Canada (NSERC) Undergraduate Student Research Award (USRA). We acknowledge the support of the Alberta Machine Intelligence Institute (Amii). 


\small
\bibliographystyle{named}
\bibliography{main.bib}

\end{document}